\documentstyle[12pt,here,mkstyle,twoside]{article}
\title{Evidence Against Evidence Theory (?!)}   
\author{Mieczys{\l}aw A. K{\l}opotek and Andrzej Matuszewski}

\newcommand{\Bem}[1]{}

\newcommand{\Bemerkung}[1]{}

\renewcommand{\baselinestretch}{1.6}
\newcommand{\rTab}[1]{Table \ref{#1}} 

\font\gh=eufm10 scaled \magstep1

\newcommand{\Prob}[2]{{ {\mbox{\gh Prob} ^{#2(#1)}} 
                         \atop {_{#1}} 
                     }}
\date{}
\begin{document}

\machetitel

\Bem{
\newpage

\thispagestyle{empty}

\quad

\newpage


\thispagestyle{empty}
}

\begin{center}
ABSTRACT - STRESZCZENIE
\end{center}

This paper is concerned with the apparent greatest weakness of the
Mathematical Theory of Evidence (MTE) of Shafer \cite{Shafer:76}, which has
been strongly criticized by Wasserman \cite{Wasserman:92ijar} - the
relationship to frequencies. \\
Weaknesses of various  proposals                  of probabilistic
interpretation of MTE belief functions  are demonstrated. \\
A new frequency-based interpretation is presented overcoming various drawbacks
of earlier interpretations.\\

\Bem{
\begin{center}
EWIDENCJA PRZECIW TEORII EWIDENCJI ?!      
\end{center}

W artykule przedstawiono zasadnicze wady znanych interpretacji cz\c{e}sto\'{s}ciowych
teorii Dempstera-Shafera. Pokazano,    \.{z}e s\c{a} one albo sprzeczne ze zdrowym
rozs\c{a}dkiem albo z regu{\l}\c{a} Dempstera sk{\l}adania niezale\.{z}nych przekona\'{n}.
Proponuje si\c{e} now\c{a} interpretacj\c{e} cz\c{e}sto\'{s}ciow\c{a} pokonuj\c{a}c w ten spos\'{o}b trudno\'{s}ci
wcze\'{s}niejszych interpretacji tej teorii.\\

\newpage


\thispagestyle{empty}

\quad

\newpage

}
 
\section{Introduction}

Wasserman in    \cite{Wasserman:92ijar}  raised  serious  concerns 
against
the Mathematical Theory of Evidence (MTE) developed by Dempster and Shafer
since 1967 - hence also called Dempster-Shafer-Theory (DST) (see 
\cite{Shafer:90ijar} for a thorough review of this theory, for major formal
definitions see Appendix A). 
One of arguments against MTE is related to  Shafer's attitude towards
frequencies. 
Shafer in  \cite{Shafer:90ijar} claims that probability theory developed over
last years from the old-style frequencies towards modern subjective
probability theory within the framework of Bayesian theory. By analogy he
claims that the very attempt to consider relation between MTE and frequencies
is old-fashioned and out of date and should be at least forbidden - for the
sake of progress of humanity. 
Wasserman opposes this view (\cite{Wasserman:92ijar}, p.371)
reminding  "major
 success story in Bayesian theory",  the exchangeability theory of 
de Finetti \cite{deFinetti:64}. It treats frequencies as special case of
Bayesian belief.  "The Bayesian 
theory contains within it a definition of frequency probability and a 
description of the exact assumptions necessary to invoke that 
definition" \cite{Wasserman:92ijar}. Wasserman dismisses Shafer's suggestion
that probability relies on analogy of frequency. .

Shafer, on the other hand, lets frequencies live a separate life. MTE beliefs
and frequencies are separated. But in this way we are left without a
definition of frequentistic belief function \cite{Wasserman:92ijar}. 

This paper is devoted to discussion of drawbacks of various interpretations of
MTE. A way out is proposed in a separate paper, the summary of our proposal is
contained in Appendix B.

\section{Basic Problems with Frequencies in MTE}

Shafer in  \cite{Shafer:90}, \cite{Shafer:90b} gave the following formal
probabilistic interpretation of belief function: 
Let $Pr$ be a probabilistic measure over the 
infinite discrete
sample space  $\Omega$, 
let   $\Gamma$ be a function  $\Gamma:\Omega \rightarrow 2^\Xi$. Then  $Bel$ 
over the space (frame of discernment) $\Xi$ is given as:
$$Bel(A) = Pr(\{\omega \in \Omega | \Gamma(\omega) \subseteq A \})$$ 
Then clearly      
$$m  (A) = Pr(\{\omega \in \Omega | \Gamma(\omega)  =        A \})$$ 
and  
$$Pl (A) = Pr(\{\omega \in \Omega | \Gamma(\omega) \cap A 
\ne \emptyset\})$$ 

Let us consider the database in \rTab{teins}.

\begin{table}[H]
\centering
\caption{Example of $\Gamma$ function}
\label{teins}
\begin{tabular}{r|r|r|r}
No.  & A & D & $\Gamma$ \\
\hline
1 & $a_1$ & $d_1$ & $\{d_1     \}$ \\
2 & $a_2$ & $d_2$ & $\{d_2,d_3 \}$ \\
3 & $a_2$ & $d_3$ & $\{d_2,d_3 \}$ \\
4 & $a_3$ & $d_3$ & $\{d_3     \}$ \\
5 & $a_4$ & $d_1$ & $\{d_1     \}$ \\
\end{tabular}
\end{table}
Let the measurable A take values  $a_1,a_2,a_3,a_4$, and let the
non-observable attribute D take values  $d_1,d_2,d_3$.
Let us define the function $\Gamma$ as capability to predict values of
attribute D given A and let us calculate it based on training sample contained
in  \rTab{teins}. We see that if A takes value 
 $a_1$, then we know that D takes value  $d_1$ - hence 
$\Gamma(A=a_1)=\{d_1\}$. Similarly values $a_3$ and  $a_4$ of attribute A 
determine uniquely the value of attribute D. But in case of 
$A=a_2$ we have an ambiguity: D is equal either  $d_2$ or  $d_3$.
Hence  $\Gamma(A=a_1)=\{d_2,d_3\}$. Now, assuming frequency probabilities
from  \rTab{teins} we calculate easily from Shafer's formula:\\
\begin{tabular}{ccc}
$m(\{d_1\})=0.4  $     &    $Bel(\{d_1\})=0.4$ &  $Pl(\{d_1\})=0.4$  \\
$m(\{d_2\})=0    $     &    $Bel(\{d_2\})=0  $ &  $Pl(\{d_2\})=0.4$  \\
$m(\{d_3\})=0.2  $     &    $Bel(\{d_3\})=0.2$ & $Pl(\{d_3\})=0.6$  \\
$m(\{d_1,d_2\})=0    $ & $Bel(\{d_1,d_2    \})=0.4$&$Pl(\{d_1,d_2  \})=0.8$\\
$m(\{d_1,d_3\})=0    $ & $Bel(\{d_1,    d_3\})=0.6$&$Pl(\{d_1,    d_3\})=1 $\\
$m(\{d_2,d_3\})=0.4  $ & $Bel(\{    d_2,d_3\})=0.6$&$Pl(\{   d_2,d_3\})=0.6$\\
$m(\{d_1,d_2,d_3\})=0$ & $Bel(\{d_1,d_2,d_3\})=1  $&$Pl(\{d_1,d_2,d_3\})=1 $\\
\end{tabular}

In probability theory two variables are independent if 
$Pr(A \cap B)=Pr(A) \cdot Pr(B)$. Let us consider two measurables A and B 
from \rTab{tzwei}. 

\begin{table}
\centering
\caption{Example of $\Gamma'$ function for a variable  B
independent of A} \label{tzwei}
\begin{tabular}{r|r|r|r|r|r}
No.  & A & B & D & $\Gamma$ & $\Gamma'$ \\
\hline
 1 & $a_1$ & $b_1$ & $d_1$ & $\{d_1     \}$ & $\{d_1,d_2,d_3 \}$ \\
 2 & $a_2$ & $b_1$ & $d_2$ & $\{d_2,d_3 \}$ & $\{d_1,d_2,d_3 \}$ \\
 3 & $a_2$ & $b_1$ & $d_3$ & $\{d_2,d_3 \}$ & $\{d_1,d_2,d_3 \}$ \\
 4 & $a_3$ & $b_1$ & $d_3$ & $\{d_3     \}$ & $\{d_1,d_2,d_3 \}$ \\
 5 & $a_4$ & $b_1$ & $d_1$ & $\{d_1     \}$ & $\{d_1,d_2,d_3 \}$ \\
 6 & $a_1$ & $b_2$ & $d_1$ & $\{d_1     \}$ & $\{d_1,d_2,d_3 \}$ \\
 7 & $a_2$ & $b_2$ & $d_2$ & $\{d_2,d_3 \}$ & $\{d_1,d_2,d_3 \}$ \\
 8 & $a_2$ & $b_2$ & $d_3$ & $\{d_2,d_3 \}$ & $\{d_1,d_2,d_3 \}$ \\
 9 & $a_3$ & $b_2$ & $d_3$ & $\{d_3     \}$ & $\{d_1,d_2,d_3 \}$ \\
10 & $a_4$ & $b_2$ & $d_1$ & $\{d_1     \}$ & $\{d_1,d_2,d_3 \}$ \\
11 & $a_1$ & $b_3$ & $d_1$ & $\{d_1     \}$ & $\{d_1,d_2,d_3 \}$ \\
12 & $a_2$ & $b_3$ & $d_2$ & $\{d_2,d_3 \}$ & $\{d_1,d_2,d_3 \}$ \\
13 & $a_2$ & $b_3$ & $d_3$ & $\{d_2,d_3 \}$ & $\{d_1,d_2,d_3 \}$ \\
14 & $a_3$ & $b_3$ & $d_3$ & $\{d_3     \}$ & $\{d_1,d_2,d_3 \}$ \\
15 & $a_4$ & $b_3$ & $d_1$ & $\{d_1     \}$ & $\{d_1,d_2,d_3 \}$ \\
16 & $a_1$ & $b_4$ & $d_1$ & $\{d_1     \}$ & $\{d_1,d_2,d_3 \}$ \\
17 & $a_2$ & $b_4$ & $d_2$ & $\{d_2,d_3 \}$ & $\{d_1,d_2,d_3 \}$ \\
18 & $a_2$ & $b_4$ & $d_3$ & $\{d_2,d_3 \}$ & $\{d_1,d_2,d_3 \}$ \\
19 & $a_3$ & $b_4$ & $d_3$ & $\{d_3     \}$ & $\{d_1,d_2,d_3 \}$ \\
20 & $a_4$ & $b_4$ & $d_1$ & $\{d_1     \}$ & $\{d_1,d_2,d_3 \}$ \\
\end{tabular}
\end{table}

Function $\Gamma$ be, as previously, be prediction of value of variable D
based on value of A, and  $\Gamma'$ be prediction of variable D given value of
 B. Let us define 
$$Bel(Z) = Pr(\{\omega \in \Omega | \Gamma(\omega) \subseteq Z \})$$ 
$$Bel'(Z) = Pr(\{\omega \in \Omega | \Gamma'(\omega) \subseteq Z \})$$ 

Let us imagine that we want to combine information from attributes A and B to
improve prediction of D by formulating a new function  $\Gamma"$ being the
base for a new belief function:
$$Bel"(Z) = Pr(\{\omega \in \Omega | \Gamma"(\omega) \subseteq Z \})$$ 

As observations being basis of functions  $\Gamma$  and 
 $\Gamma'$ are obviously independent, so one would expect that the belief
function  $Bel"$ is simply the combination OF INDEPENDENT EVIDENCE 
$Bel$  and  $Bel'$ via Dempster rule.  And this is in fact the case:
$$Bel"=Bel \oplus Bel'$$ 
But there is one weak point in all of this: $Bel'$ is (and will always be) a
vacuous belief function, hence it does not contribute anything to our
knowledge of the value of the attribute. Reverting this example we can say
that whenever we combine two non-vacuous belief functions, then the
measurements underlying their empirical calculation are for sure statistically
dependent. So we claim that:\\
{\em Under Shafer's frequentist interpretation, if two belief functions are
(statistically) independent then at least one of them is non-informative.} 

Another practical limitation of Shafer's probabilistic interpretation is 
consideration of conditional beliefs. 
Let as look at \rTab{tdrei}. 

\begin{table}
\centering
\caption{Scheme of creation of conditional probability}
\label{tdrei}
\begin{tabular}{r|r|r|r}
No.  & A & $A=a_1 \lor A=a_2$ & $A|A=a_1 \lor A=a_2$ \\
\hline
1 & $a_1$ & yes& $a_1$ \\
2 & $a_2$ & yes& $a_2$ \\
3 & $a_2$ & yes& $a_2$ \\
4 & $a_3$ & no & $ - $ \\
5 & $a_4$ & no & $ - $  \\
\end{tabular}
\end{table}

If we want to calculate conditional probability of  $A=a_1$ given         
observation that A takes only one of values $a_1$ or  $a_2$, 
we select cases from the database fitting the condition   $A=a_1 
\lor A=a_2$, and thereafter within this subset we calculate frequency
probabilities:  $Pr(A=a_1|A=a_1 \lor A=a_2)=1/3=0.33$.
Now, based on \rTab{tvier} let us run similar procedure for MTE beliefs.

\begin{table}
\centering
\caption{Example of conditioning in MTE}
\label{tvier}
\begin{tabular}{r|r|r|r|r|r}
No.  & A & D & $\Gamma$ & $\Gamma \cap\{d_1,d_2 \}$ & $\Gamma'=$\\
     &   &   &          & $\neq \emptyset$   & $\Gamma \cap\{d_1,d_2 \}$\\
\hline
1 & $a_1$ & $d_1$ & $\{d_1     \}$ & yes & $\{d_1     \}$ \\
2 & $a_2$ & $d_2$ & $\{d_2,d_3 \}$ & yes & $\{d_2     \}$  \\
3 & $a_2$ & $d_3$ & $\{d_2,d_3 \}$ & yes & $\{d_2     \}$  \\
4 & $a_3$ & $d_3$ & $\{d_3     \}$ & no  & $      -     $  \\
5 & $a_4$ & $d_1$ & $\{d_1     \}$ & no  & $      -     $   \\
\end{tabular}
\end{table}

Let us assume that we want to find out our degree of belief in values of D
given that only values  $d_1$ or $d_2$ are allowed.  For this purpose
we restrict the set of cases to those cases  $\Omega'$ for which our function
 $\Gamma$ has non-empty intersection with the set of values of interest. 
For this group of cases we define the function 
 $\Gamma'(\omega)= \Gamma(\omega) \cap \{d_1,d_2
\}$. Let :\\
$$Bel(Z) = Pr(\{\omega \in \Omega | \Gamma(\omega) \subseteq Z \})$$ 
$$Bel'(Z) = Pr(\{\omega \in \Omega' | \Gamma'(\omega) \subseteq Z \})$$ 

Additionally let us define the simple support function $Bel"$ such that
$m"(\{d_1,d_2\})=1$. It is easily seen that:\\
$$Bel' = Bel \oplus Bel"$$
(as expected because the expression  $Bel \oplus Bel"$
means shaferian conditioning on event
$\{d_1,d_2\}$). 
And everything would be O.K. if it were not that the function 
$\Gamma'$ has little to do with the non-observable attribute $D$ - compare
line no.3 of \rTab{tvier}. 
Let us remind that function $\Gamma$ 
represented by definition for a given observed value of variable A
the set of potentially possible values of attribute D, deducible from the
training sample. For every object $\omega$, if we know the true value 
$a$ of $A$ one of the values from the set $\Gamma(\omega)$ was the true value
of
D for this object $\omega$. But within  $\Gamma'(\omega)$ the true value of
attribute D does not need to be contained - compare
line no.3 of \rTab{tvier}. 
But, let us remind, Shafer claimed 
\cite{Shafer:90,Shafer:90b} that function  $\Gamma$ indicates that
the variable takes for object 
 $\omega$ one of the values  $\Gamma(\omega)$. But we have just demonstrated
that already after a single step of conditioning function  $\Gamma'$
simply tells lies. Its meaning is not dependent solely on subpopulation
$\Omega'$, to which it refers, but also on the history, how this population
was selected. But we had for probability distributions that after conditioning
a variable for not rejected objects took always those values which were
indicated by the result of conditioning.

Both above failures of Shafer's probabilistic interpretation of his own
theory of evidence were driving forces behind the elaboration of a new
probabilistic interpretation of MTE presented subsequently. \\

We shall summarize this section saying that: {\em Shafer's probabilistic
interpretation of Dempster's \& Shafer's Mathematical Theory of Evidence is
not compatible with this theory: It does not fit the Dempster's rule of
combination of independent evidence}. .

As statistical properties of Shafer's \cite{Shafer:76} notion of evidence are 
concerned, further    criticism has been expressed by Halpern and Fagin 
(\cite{Halpern:92} in sections 4-5). Essentially 
the criticism is
pointed there at the fact that "the belief that represents the joint 
observation is equal to the combination is in general not equal to the 
combination of the belief functions representing the individual (independent) 
observations" (p.297). The other point raised there that though it is possible
to capture properly in belief functions evidence in terms of 
probability  of observations update functions (section 4 of 
\cite{Halpern:92}), it is not possible to do the same if we would like to 
capture evidence in terms of beliefs  of observations update functions  
(section 5 of \cite{Halpern:92}). 

\section{Smets' Approach to Frequencies}

 Smets \cite{Smets:92} has made some strong 
statements in defense of the Dempster-Shafer theory against sharp criticism 
of this theory by its opponents as well as unfortunate users of the MTE who 
wanted to attach it to the "dirty reality" (that is objectively given 
databases). He 
insisted on Bels  not being connected to any empirical measure (frequency, 
probability 
etc.) considering the domain of MTE applications  as the one where 
"we are ignorant of the existence of probabilities", and not one with 
         "poorly known probabilities" (\cite{Smets:92}, p.324). 
The basic property of probability, which should be dropped in the MTE 
axiomatization, should be  the additivity of belief measures.  
Surely, it is easy  to imagine situations where - in the real life - the 
additivity is not granted: 
Imagine we have had a cage with 3 pigs, we put into it 3 
hungry lions two hours ago, how many animals are there now ? ($3+3 <6$). Or 
ten years ago we left one young man and one young woman on an island in the 
 middle of the atlantic ocean with food and weapons sufficing for 20 years. 
How many human beings are there now ? ($1+1>2$). \\
The trouble is, however, that the objects stored in databases of a computer 
behave usually (under normal operation) in an additive manner. Hence the MTE 
is simply disqualified for any reasoning within human collected data on real 
world, if we accept the philosophy of Smets and Shafer. 

The question may be raised at this point, what else practically useful 
can be obtained 
from a computer reasoning on the basis of such a MTE. 
If the MTE models, as Smets and Shafer claim, human behaviour during 
evidential reasoning, then  
it would have to be 
demonstrated that humans indeed reason as MTE. 
We take e.g. 1000 people who never heard of Dempster-Shafer theory,
briefly explain the static component, provide them with two opinions of 
 independent experts and expect of them to answers what are their final 
beliefs.
Should their answers correspond to results of the MTE (at least converge 
toward them), then the computer, if fed 
with our knowledge, would 
be capable to predict our conclusions on a given subject. However, to our 
knowledge, no experiment like this has ever been carried out. 
 Under these circumstances 
the computer reasoning with MTE would tell us what we have to think and 
not 
what we think. But we don't suspect that anybody would be happy about a 
computer like this.

Hence, from the point of view of computer implementation the 
philosophy of Smets and Shafer is not acceptable.
Compare also Discussion in \cite{Halpern:92} on the subject.

Smets        felt a bit uneasy about a total loss of reference to any 
scientific experiment checking practical applicability of the MTE and 
suggested some probabilistic background for decision making (e.g. the 
pigeonistic probabilities of Smets), but we are afraid that by these 
interpretations he falls  precisely into the same pitfalls  he  claimed to 
avoid by his   highly abstract philosophy.

As Smets probabilistic interpretations are concerned, let us "continue" the 
killer example of \cite{Smets:92} on pages 330-331. "There are three 
potential killers, A, B, C. Each can use a gun or a knife. We shall select
one of them, but you will not know how we select the killer. The killer
selects 
 his weapon by a random process with p(gun)=0.2 and p(knife)=0.8. Each of A, 
B, C 
has his own personal random device, the random devices are unrelated. ...... 
Suppose you are a Bayesian and you must express your "belief" that the killer 
will use a gun. The BF (belief function) solution gives $Bel(gun)=0.2 \times 
0.2 \times 0.2=0.008$. ..... Would you defend 0.2 ? But this applies only if I
select a killer with a random device ...... But we never said we would use a 
random device; we might be a very hostile player and cheat whenever we can.
...
. So you could interpret Bel(x) as the probability that you are sure to win 
whatever Mother Nature (however hostile) will do." \\
Yes, we will try to continue the hostile Mother Nature game here. For 
completeness we understand that $Bel(knife)=0.8 ^3=0.512$ and 
$Bel(\{gun,knife\})=1$. But suppose there is another I, the chief of gangster
science fiction 
physicians, making decisions independly of the chief I of the killers. The 
chief I of physicians knows of the planned murder and has three physicians 
X,Y,Z.  Each can either rescue a killed man or let him die. 
I shall select one 
of them, but you will not know how I select the physician. The physician, in 
case of killing with a gun, selects 
his attitude by a random process with $p(rescue|gun)=0.2$ 
and $p(let\  die|gun)=0.8$ and he lets the person die otherwise. Each 
of X, Y, Z 
has his own personal random device, the random devices are unrelated. ...... 
Suppose you are a Bayesian and you must express your "belief" that the 
physician will rescue if the killer 
will use a gun. The BF (belief function) solution gives 
$Bel_1(rescue|gun)=0.2^3=0.008$. $Bel_1(let\  die|gun)=0.8^3=0.512$, 
$Bel_1(\{recue,let \  die\}|gun)=1$. Also  $Bel_2(let\  
die|knife)=1$. As the scenarios for $Bel_1$ and $Bel_2$ are independent, let 
us combine them by the Dempster rule: $Bel_{12}=Bel_1 \oplus Bel_2$. We make 
use of the Smets' claim that "the de re and de dicto interpretations lead to 
the same results" (\cite{Smets:92}, p. 333), that is $Bel(A|B)=Bel(\lnot B 
\lor A)$. Hence 
$$m_{12}(\{(gun,let\ die),(knife,let\ die),(gun  ,rescue)\})=0.480$$
$$m_{12}(\{(gun,rescue),(knife,let\ die)\})=0.008$$
$$m_{12}(\{(knife,let\ die),(gun,let\ die)\})=0.512$$

Now let us combine $Bel_{12}$ with the original $Bel$. We obtain:\\
$$m\oplus m_{12}((gun,let\ die)=0.008 \cdot 0.480+0.008 \cdot 0.512=
0.008 \cdot 0.992$$

But these two unfriendly chiefs of gangster organizations can be extremely 
unfriendly and in fact your chance of winning a bet may be as bad as $0.008 
\cdot 0.512$ for the event $(gun,let\ die)$. Hence the "model" proposed by 
Smets for understanding beliefs functions as "unfriendly Mother Nature" is 
simply wrong. If the Reader finds the combination of $Bel_2$ with the other 
Bels a little tricky, then for justification He should refer to the paper of 
Smets and have a closer look at all the other examples. \\

Now returning to the philosophy of "subjectivity" of Bel measures:
Even if a human being may 
possess his private view on a subject, it is only after we formalize the 
feeling of subjectiveness and hence ground it in the data that we can rely on 
 any computer's "opinion". We hope we have found one such formalization in 
this 
paper. The notion of labeling developed here substitutes one aspect of 
subjective human behaviour - if one has found one plausible explanation, one 
is too lazy to look for another one. So the process of labeling may express 
our personal attitudes, prejudices, sympathies etc. The interpretation drops 
deliberately the strive for maximal objectiveness aimed at by traditional 
statistical analysis. Hence we think this may be a promising path for further 
research going beyond the DS-Theory formalism.

Smets \cite{Smets:92} views  the probability theory as a formal mathematical 
apparatus and hence puts it on the same footing as his view of the MTE. 
However, in our opinion, he ignores totally one important thing: The abstract 
concept of probability has its real world counterpart of relative frequency 
which tends to behave approximately like the theoretical probability in 
sufficiently many experimental settings as to make the abstract concept of 
probability useful for practical life. And a man-in-the-street will expect of 
the MTE to possess also such a counterpart or otherwise the MTE will be 
 considered as another version of the theory of counting devils on a 
pin-head.\\ 

It is worth mentioning that Smets made recently an attempt to justify usage
of belief functions instead of Bayesian probabilities, to identify situations
in which usage of belief functions is more reasonable than usage of
probabilities \cite{Smets:94}. However, the data modifying impact of
application of Dempster-rule is not explicitly recognized there and numerical
examples presented there deliberately avoid situations where more than one
belief function may have more than one focal point.\\

\section{MTE and Random Sets}

The canonic 
random set 
interpretation 
\cite{Nguyen:78}
is one with a statistical process over set
instantiations. The rule of combination assumes then that two such 
statistically independent processes are run and we are interested in their 
intersections. This approach is not sound as empty intersection is excluded  
and this will render any two processes statistically dependent.
We overcome this difficulty assuming in a straightforward manner that 
we are "walking" from population to population applying the Rule of 
Combination. Classical DS theory in fact assumes such a walk implicitly
or it drops in fact the assumption that Bel() of the empty set is equal 0.
In this sense the random set approaches may be considered as sound as ours. 

However, in many cases the applications of the model are insane. 
 For example, 
to imitate the logical inference it is frequently assumed that we have a 
Bel-function describing the actual observed value of a predicate P(x), 
 and a Bel-function describing the implication "If P(x) then Q(x)"
\cite{Ma:91}. It is 
assumed further that the evidence on the validity of both Bel's has been 
collected independently and one applies the DS-rule of combination to 
calculate the Bel of the predicate Q(x). One has then to assume that there is 
a focal m of the following expression: $m(\{(P(x) , Q(x)),
(\lnot P(x) , Q(x)),(\lnot P(x) ,\lnot  Q(x)) \})$ which actually means that 
with non-zero probability at the same time $P(x)$ and $\lnot P(x)$ hold for 
 the same object as we will see in the following example:  
Let $Bel_1$ represent our belief in the implication, with focal points:
$$m_1(P(x) \rightarrow Q(x))=0.5, 
\ m_1(\lnot (P(x) \rightarrow Q(x)))=0.5, $$
Let further the independent opinion $Bel_2$ on P(x) be available in the form 
of focal points:
$$m_2(P(x))=0.5, \ m_2(\lnot P(x))=0.5$$
Let $Bel_{12}=Bel_1 \oplus Bel_2$  represent the combined opinions of both 
experts. The focal points of $Bel_{12}$ are:
$$ m_{12}(\{(P(x) , Q(x))\})=0.33, \ 
   m_{12}(\{(P(x) ,\lnot Q(x))\})=0.33, $$  
$$   m_{12}(\{(\lnot P(x) , Q(x)),(\lnot P(x) ,\lnot  Q(x)) \})=0.33$$ 

$m_{12}(\{(P(x) , Q(x))\})=0.33$  makes us believe that there exist objects 
for 
which both P(x) and Q(x) holds. However, a sober (statistical) look at expert 
opinions suggests that  all situations for which the implication $P(x) 
\rightarrow Q(x)$ holds, must result from falsity of $P(x)$, hence whenever 
Q(x) holds then $\lnot P(x)$ holds. These two facts combined mean that P(x) 
and its negation have to hold simultaneously. 
 This is actually absurdity overseen deliberately. The source of this 
 misunderstanding is obvious: the lack of proper definition of what is and 
what is  not independent. 
 Our interpretation allows 
for 
sanitation of this situation. We are not telling that the predicate and its 
negation hold simultaneously. Instead we say that
for one object we modify 
 the measurement procedure (set a label) in such a way that it,
applied for calculation of $P(x)$, yields true and at the same time 
for another object, with the same original properties 
we make another modification of 
measurement procedure (attach a label to it) so that 
 measurement of $\lnot P(x)$ yields also 
true, because possibly two different persons were enforcing their different 
beliefs onto different subsets of data.\\

Our approach is also superior to canonical random set approach in the 
following sense: The 
canonical approach requires knowledge of the complete  random  set 
realizations 
of two processes on an object to determine the combination of both processes. 
 We, however, postpone the acquisition of
 knowledge of the precise 
instantiation of properties 
of the object by interleaving the concept of measurement and the concept of 
labeling process. This has a close resemblance to practical processing 
whenever diagnosis for a patient is made. If a physician finds a set of 
hypotheses explaining the symptoms of a patient, he will 
usually not
 try to carry 
out other testing procedures than those related to the plausible hypotheses.
He runs clearly at risk that there exists a different set of hypotheses
also explaining the patients's symptoms, and so a disease unit possibly 
present may not be detected on time, but usually the risk is sufficiently
low to proceed in this way, and the cost savings may prove enormous. \\

\section{Upper and Lower Probabilities}

Still another approach was to handle Bel and Pl as lower and upper 
probabilities
 \cite{Dempster:67}. This approach is of limited use as not every set of 
lower and upper probabilities leads to Bel/Pl functions \cite{Kyburg:87},
hence establishing a unidirectional relationship between probability theory 
and the DS-theory. Under our interpretation, 
the Bel/Pl function pair may be considered as a kind of interval 
approximations to some "intrinsic" probability distributions which, however, 
cannot be accessed by feasible  measurements and are only of interest as a 
kind 
of qualitative explanation to the physical quantities really measured.\\

Therefore another approach was to handle them as lower/upper envelops to some 
probability  function realization  \cite{Kyburg:87},  
  \cite{Fagin:91B}. However, the DS rule of combination of independent 
evidence 
failed. \\

\section{Inner and Outer Measures}

Still another approach was to handle Bels/Pl in probabilistic structures 
rather than in probabilistic spaces \cite{Fagin:91}. Here, DS-rule could be 
justified as 
one of the possible outcomes of independent combinations, but no stronger 
properties were available. This is due to the previously mentioned fact that 
exclusion of empty intersections renders actually most of conceivable 
processes dependent. Please notice that under our interpretation no 
such ambiguity occurs. This is because we not only drop empty intersecting 
 objects but also relabel the remaining ones so that any probability 
calculated 
afterwards does not refer to the original population.

So it was tried to drop the DS-rule altogether in the probabilistic 
structures, but then it was not possible to find a meaningful rule for 
multistage  reasoning \cite{Halpern:92}. This is a very important negative 
outcome. As the Dempster-Shafer-Theory is sound in this respect and possesses 
many  useful properties (as mentioned in the Introduction), it should be 
sought 
for an interpretation meeting the axiomatic system of DS Theory rather then 
tried to violate its fundamentals. Hence we consider our interpretation as a 
promising one for which decomposition of the joint distribution paralleling 
the results for probability distributions may be found based on the data.\\

\section{Rough Set Approach}

An interesting alternative interpretation of the Dempster-Shafer Theory was 
found within the framework of the rough set theory \cite{Skowron:93}, 
\cite{Grzymala:91}. Essentially the rough set theory searches for 
approximation of the value of a decision attribute by some other
(explaining) attributes. 
 It usually happens that those attributes are capable only of providing a 
lower 
and upper approximation to the value of the decision attribute (that is the 
set of vectors of explaining attributes 
supporting only this value 
 of the decision variable,
and the set of vectors of explaining attributes 
supporting also this value 
 of the decision variable resp.- for details 
see texts of Skowron \cite{Skowron:93} and Grzyma{\l}a-Busse 
\cite{Grzymala:91}).
The Dempster Rule of combination is interpreted by Skowron \cite{Skowron:93b} 
as combination of opinions of independent experts, who possibly look at 
different sets of explanation attributes and hence may propose different 
explanations. 

The difference between  our approach and the one based on rough sets lies 
first of all in the "ideological" background: We assume that the "decision 
attribute" is set-valued whereas the rough-set approach assumes it to be 
single-valued. This could have been overcome by some tricks which will not be 
explained in detail here.
But the combination step is here essential: If 
we assume that the data sets for forming  knowledge of these two experts are 
exhaustive, then it can never occur that these opinions are contradictory. 
 But 
the MTE rule of combination uses the normalization factor for dealing with 
cases like this.
Also the opinions of experts may have only the form of a simple (that is 
deterministic) support function. Hence, rough-set interpretation implies 
axioms not actually present in the MTE. Hence rough set interpretation is 
 on the one hand restrictive, and on the other hand not fully conforming to 
the general MTE. From our point of view the MTE would change the values of 
decision variables rather then recover them from expert opinions.

Here, we come again at the problem of viewing the independence of experts.  
The MTE assumes some strange kind of independence within the data: the 
proportionality of the distribution of masses of sets of values among 
intersecting subsets weight by their masses in the other expert opinion.  
Particularly unfortune is the fact for the rough set theory, that given a 
value of the decision variable, the respective indicating vectors of 
explaining variables values must be proportionally distributed among the 
experts not only for this decision attribute value, but also for all the 
other decision attribute values that ever belong to the same focal point.  
Hence applicability of the rough set approach is hard to justify by a 
simple(, "usual"  as Shafer wants) statistical test. 
On the other hand, statistical independence required for Dempster rule 
application within our approach can be easily checked.

To demonstrate the problem of rough set theory with  combination of opinions
of independent experts let us consider an examle of two experts having the 
combined explanatory attributes $E_1$ (for expert 1) and $E_2$ (for expert 2) 
both trying to guess the decision attribute $D$. Let us assume that $D$ takes 
one of two values: $d_1,d_2$, $E_1$ takes one of three values $e_{11}, e_{12},
e_{13}$, $E_2$ takes one of three values $e_{21}, e_{22}, 
e_{23}$. Furthermore let us assume that the rough set analysis
of an exhaustive set of possible cases
 shows that the 
value $e_{11}$ of the attribute $E_1$ indicates the value $d_1$ of the 
decision attribute $D$, 
$e_{12}$ indicates $d_2$,
$e_{13}$ indicates the set \{$d_1,d_2$\},
 Also        let us assume that the rough set analysis of an exhaustive set of
possible cases shows that the 
value $e_{21}$ of the attribute $E_2$ indicates the value $d_1$ of the 
decision attribute $D$, 
$e_{22}$ indicates $d_2$,
$e_{32}$ indicates the set \{$d_1,d_2$\},
 From the point of view of 
Bayesian analysis four cases of causal influence may be distinguished (arrows 
indicate the direction of dependence).
$$E_1 \rightarrow D \rightarrow  E_2$$
$$E_1 \leftarrow  D \leftarrow   E_2$$
$$E_1 \leftarrow  D \rightarrow  E_2$$
$$E_1 \rightarrow D \leftarrow   E_2$$

 From the point of view of Bayesian analysis, in the last case attributes 
$E_1$ and $E_2$ have to be unconditionally independent, in the remaining 
cases: $E_1$ and $E_2$ have to be independent  conditioned on $D$. 
Let us consider first unconditional independence of $E_1$ and $E_2$. Then 
we have that  (For meaning of $\Prob{\omega}{P}$ see
Appendix B):
$$(\Prob{\omega}{P} E_1(\omega)=e_{11} \land E_2(\omega)=e_{22}) = $$
$$=
(\Prob{\omega}{P} E_1(\omega)=e_{11} ) \cdot 
(\Prob{\omega}{P}  E_2(\omega)=e_{22}) >0  $$
 However, it is impossible that $(\Prob{\omega}{P} E_1(\omega)=e_{11} \land 
E_2(\omega)=e_{22}) > 0$  because we have to do with experts who may provide 
us possibly with information not specific enough, but will never provide us 
with 
contradictory information. We conclude that unconditional independence of 
experts is impossible.\\
Let us turn to independence of $E_1$ and $E_2$ if  conditioned on $D$. 
We introduce the following denotation:\\
$$p_1 = \Prob{\omega}{P} D(\omega)=d_1$$
$$p_2 = \Prob{\omega}{P} D(\omega)=d_2$$
$$e_1'= \Prob{\omega}{ (D(\omega)=d_1)\land P} E_1(\omega)=e_{11}$$
$$e_3'= \Prob{\omega}{ (D(\omega)=d_1)\land P} E_1(\omega)=e_{13}$$
$$f_1'= \Prob{\omega}{ (D(\omega)=d_1)\land P} E_2(\omega)=e_{21}$$
$$f_3'= \Prob{\omega}{ (D(\omega)=d_1)\land P} E_2(\omega)=e_{23}$$
$$e_2"= \Prob{\omega}{ (D(\omega)=d_2)\land P} E_1(\omega)=e_{12}$$
$$e_3"= \Prob{\omega}{ (D(\omega)=d_2)\land P} E_1(\omega)=e_{13}$$
$$f_2"= \Prob{\omega}{ (D(\omega)=d_2)\land P} E_2(\omega)=e_{22}$$
$$f_3"= \Prob{\omega}{ (D(\omega)=d_2)\land P} E_2(\omega)=e_{23}$$
 Let $Bel_1$ and $m_1$ be the belief function and the mass function 
representing the knowledge of the first expert, let $Bel_2$ and $m_2$ be the 
belief function and the mass function 
representing the knowledge of the second expert. Let $Bel_{12}$ and $m_{12}$ 
be the belief function and the mass function 
representing the knowledge contained in the combined usage of attributes 
$E_1,E_2$ if used for prediction of $D$ - on the grounds of the rough set 
theory. It can be easily checked that:\\
$$m_1(\{d_1\})=e_1' \cdot p_1,\  m_1(\{d_2\})=e_2" \cdot p_2,\ 
m_1(\{d_1,d_2\})=e_3' \cdot p_1, + e_3"' \cdot p_2$$
$$m_2(\{d_1\})=f_1' \cdot p_1,\  m_2(\{d_2\})=f_2" \cdot p_2,\ 
m_2(\{d_1,d_2\})=f_3' \cdot p_1, + f_3"' \cdot p_2$$
and if we assume the conditional independence of $E_1$ and $E_2$ conditioned 
on $D$, then we obtain:
$$m_{12}(\{d_1\})=e_1' \cdot f_1' \cdot p_1 +
e_1' \cdot f_3' \cdot p_1 +
e_3' \cdot f_1' \cdot p_1 $$
$$m_{12}(\{d_2\})=e_2" \cdot f_2" \cdot p_2 +
e_2" \cdot f_3" \cdot p_2 +
e_3" \cdot f_2" \cdot p_2 $$
$$m_{12}(\{d_1,d_2\})=e_3' \cdot f_3' \cdot p_1 +
e_3" \cdot f_3" \cdot p_2$$
 However, the Dempster rule of combination would result in (c - normalization 
constant):\\
$$m_1\oplus m_2(\{d_1\})=c \cdot (e_1' \cdot f_1' \cdot p_1^2 +
e_1' \cdot f_3' \cdot p_1^2         +
e_1' \cdot f_3" \cdot p_1 \cdot p_2 +
e_3' \cdot f_1' \cdot p_1^2         +
e_3" \cdot f_1' \cdot p_1 \cdot p_2)$$
$$m_1\oplus m_2(\{d_2\})=c \cdot (e_2" \cdot f_2" \cdot p_2^2 +
e_2" \cdot f_3' \cdot p_1 \cdot p_2 +
e_2" \cdot f_3" \cdot p_2^2         +
e_3' \cdot f_2" \cdot p_1 \cdot p_2 +
e_3" \cdot f_2" \cdot p_2^2        )$$
$$m_1\oplus m_2(\{d_1,d_2\})=c \cdot 
e_3' \cdot f_3' \cdot p_1^2         +
e_3" \cdot f_3" \cdot p_2^2         +
e_3' \cdot f_3" \cdot p_1 \cdot p_2 +
e_3" \cdot f_3' \cdot p_1 \cdot p_2)$$
Obviously, $Bel_{12}$ and $Bel_1\oplus Bel_2$ are not identical in general.  
We conclude that conditional independence of experts is also impossible. 
Hence no usual staatistical indeperndence assumption  is valid for the 
rough set interpretation of the MTE. This fact points at where the difference 
between rough set interpretation and our interpretation lies in: in our 
 interpretation, traditional statistical independence is incorporated into 
the Dempster's scheme of combination (labelling process).

By the way, lack of correspondence between statistical independence and 
Dempster rule of combination is characteristic not only for the rough set 
interpretation, but also of most of the other ones, e.g.
\cite{Hummel:88,Smets:93}. The 
Reader should read carefully clumsy statements of Shafer about MTE 
and statistical independence in \cite{Shafer:90b}.

\section{Probability of Provability}

Now 
we  draw attention to the way several authors are writing
about "independence of experts' opinions". Smets \cite{Smets:93}
suggests
that independence in MTE should be understood {\em intuitively} in terms of
experts independence - only intuitively, as 
was shown in the previous section.
Hummel and Landy \cite{Hummel:88}
give still more curious account of expert
independence: It is a (totally ignorant) "committee" that combines
opinions of experts which, it supposes, are independent. 

We think that, what is really meant behind those clumsy expert opinion
independence notions, has been correctly and explicitly stated is the
probability of provability approach \cite{Pearl:90,Provan:90}. The probability
of provability approach means the following: we have a deterministic state of
affairs. We have experts that make deterministic (categoric) statements about
the state of affairs, each statement stemming from a different expert. To
derive any theorem about the world we use propositional calculus. However,
we know that experts may make errors (independently of one another), that is
that their statements may be
actually wrong. So each proposition is assigned a probability of being 
correct.
What MTE achieves under this interpretation (if we refrain from normalization)
is calculation of probability that the proof of our final statement is
correct. First notice, that we do not calculate the probability that the
statement itself is correct (this probability will be usually higher). Then
imagine what space of events is considered for calculating or guessing the
expert's reliability. We are not concerned with any actual set of observations,
 but rather with the set of observations together with sets of opinions
expressend by each expert on each observation. And we are not building a world
model but rather a model of the set of experts. And to reason about a new
event we cannot rely on previous knowledge (as we do not derive it), but on
experts' opinions (which we then valuate with experience about experts'
reliability). A statistical evaluation will be rather hard under such
circumstances. First of all how an error of an expert could be treated~?
Should
be a more specific error be considered the same way as an error in a more
general statement or not~?. Last not least we lose any statistical information
about the real world behaviour under such interpretation.\\

\section{General Remarks}

The Dempster-Shafer Theory exists already over two decades. Though it was 
claimed to reflect various aspects of human reasoning, it has not been widely 
used in expert systems until recently due to the high computational 
complexity. Three years ago, however, an important paper of Shenoy and Shafer 
\cite{Shenoy:90} has been published, 
along with papers of other authors similar in spirit,
which meant a break-through for 
application of both Bayesian and Dempster-Shafer theories in  reasoning 
systems, because it demonstrated that if joint (Bayesian or DS) belief 
distribution can be decomposed in form of a belief network than it can be both
represented in a compact manner and marginalized efficiently by local 
computations. 

This  fact  makes  them  suitable   as   alternative 
fundamentals for  representation  of  (uncertain)  knowledge in  
expert system knowledge bases \cite{Henrion:90}. 

Reasoning in Bayesian belief networks has been subject of intense
research work also earlier 
 \cite{Shachter:90}, \cite{Shenoy:90}, \cite{Pearl:86},
\cite{Pearl:88}. 
There exist methods of  imposing 
various logical constraints on the probability  density  function  and  of 
calculating  marginals  not  only  of  single  variables  but  of 
complicated logical expressions over elementary statements of  the 
type X =x   (x   belonging to the domain of the variable X ) 
\cite{Pearl:88}. 
     There  exist  also  methods  determining   the 
decomposition of 
a joint probability distribution given by a sample into a 
Bayesian belief network
\cite{Chow:68}, \cite{Rebane:89}, \cite{Acid:91}.

     It  is  also  known  that  formally 
probability distributions can be treated as special cases of 
Dempster-Shafer belief distributions (with singleton focal points) 
\cite{Halpern:92}.
 
     However, for application of DS Belief-Functions
for representation of uncertainty in 
expert 
system knowledge bases there exist several severe  obstacles.   The 
main one  is  the  missing  frequentist  interpretation  of  the 
DS-Belief function and hence neither a comparison of the deduction 
results with experimental  data  nor  any  quantitative  nor  even 
qualitative conclusions can be drawn from results of deduction  in 
Dempster-Shafer-theory based expert systems \cite{Ma:91}.

     Numerous attempts to 
find  a frequentist  interpretation have been reported (e.g. 
\cite{Fagin:91}, \cite{Fagin:91B},
\cite{Grzymala:91},
\cite{Halpern:92}, 
\cite{Kyburg:87}, \cite{Shafer:90b},
 \cite{Skowron:93}).   But, as Smets \cite{Smets:92} states, they failed 
either trying 
 to incorporate  Dempster  rule  or  when  explaining  the  nature  of 
probability interval approximation. 
The Dempster-Shafer Theory experienced 
therefore
sharp criticism from several
authors in the past \cite{Pearl:88}, \cite{Halpern:92}.
It is suggested in those critical papers that 
the claim of MTE to represent uncertainty stemming from ignorance is not 
valid.    Hence alternative rules of combination of evidence have been 
proposed. However, these rules fail to fulfill Shenoy-Shafer axioms of local 
computation \cite{Shenoy:90} and hence are not tractable in practice. These 
failures of those authors meant to us that one shall nonetheless try to find 
a meaningful frequentist interpretation of MTE compatible with Dempster rule 
of combination.

     We have carefully studied several of these approaches and are 
convinced that the key for many of those  failures  (beside  those 
mentioned by Halpern in \cite{Halpern:92}) was:
(1)  treating  the  Bel-Pl 
pair as an interval approximation and
(2)  viewing combination of evidence as a process of approaching a point 
estimation. In this paper we claim 
that the most reasonable treatment of Bel's Pl's is to consider them to be 
POINT ESTIMATES of probability distribution over set-valued attributes
 (rather then interval estimates of probability distribution over single 
valued attributes). Of course, we claim 
 also that Bel-Pl estimates by  an  interval  some  probability  density 
function but in our interpretation  that  "intrinsic"  probability 
density function is of little interest for the  user.  The combination of 
evidence represents in our interpretation manipulation of data by imposing on 
them our prejudices (rather then striving for extraction of true values).

 Under  these 
assumptions a frequentionistically meaningful interpretation of the 
Bel's  can  be  constructed (Appendix B),  which   remains   consistent
under combination  of  joint  distribution   with   "evidence",   giving 
concrete  quantitative  meaning  to  results  of   expert   system 
reasoning. Within this interpretation we were  able 
to prove the correctness of Dempster-Shafer rule. This means  that 
this frequentist interpretation  is  consistent  with  the  DS-Theory  to  the
largest extent ever  achieved. 

Finally, we feel obliged to apologize and to say that all critical remarks 
towards 
interpretations of MTE elaborated by other authors result from deviations of 
those interpretations from the formalism of the MTE. We do not consider, 
 however, a deviation from MTE as a crime, because modifications of MTE may 
and possibly have a greater practical importance than the original theory. 
The purpose of this paper was to shed  a bit more light onto the intrinsic 
nature of  pure MTE and not to call for orthodox attitudes towards MTE.

\newcommand{\ReadingsIn}{G. Shafer, J. Pearl eds: Readings in Uncertain 
Reasoning, (
Morgan Kaufmann Publishers Inc., San Mateo, California, 1990)}

\small

\renewcommand{\baselinestretch}{0.8}
{

\section*{Appendix A: Formal  Definitions of MTE}

Let $\Xi$ be a finite  set of elements called elementary events. 
Any subset of $\Xi$ be a composite event. $\Xi$ be called also the 
frame of discernment.\\
A basic probability assignment function m:$2^\Xi  \rightarrow [0,1]$
such that  $$  \sum_{A \in 2^\Xi } |m(A)|=1 $$
$$  m(  \emptyset) =0 $$
$$\forall_{A \in 2^\Xi} \quad  0 \leq  \sum_{A \subseteq B} m(B)$$
($|.|$ - absolute value).\\
      
A belief function be defined as Bel:$2^\Xi \rightarrow [0,1]$ so that 
 $Bel(A) = \sum_{B \subseteq A} m(B)$
A plausibility function be Pl:$2^\Xi \rightarrow [ 0,1]$  with 
$\forall_{A \in 2^\Xi} \  Pl(A) = 1-Bel(\Xi-A )$.
A commonnality function be Q:$2^\Xi \rightarrow [0,1]$ with 
 $\forall_{A \in 2^\Xi} \quad Q(A) = \sum_{A \subseteq B} m(B)$

Furthermore, a Rule of Combination of two Independent Belief Functions 
$Bel_1$,
 $Bel_2$ Over the Same Frame of Discernment (the so-called Dempster-Rule),
denoted 
    $$Bel_{E_1,E_2}=Bel_{E_1} \oplus Bel_{E_2}$$ 
 is defined as follows: :
$$m_{E_1,E_2}(A)=c \cdot  \sum_{B,C; A= B \cap C} m_{E_1}(B) \cdot 
m_{E_2}(C)$$ (c - constant normalizing the sum of $|m|$ to 1)

 Whenever $m(A) > 0$, we say that A is the focal  point  of  the 
Bel-function.
 If the only focal point of a belief function is $\Xi$ ($m(\Xi)=1$), then 
Bel is called vacuous belief function (it does not contain any information on
whatever value is taken by the variable).

\section*{Appendix B: Review of the New Interpretation of MTE}
(A summary of the paper \cite{Klopotek:933a}).
\newcommand{\LP}{\mbox{\bf LP}}
  $\Prob{x}{P}\alpha(x)$ - the probability of  $\alpha(x)]$ being true within 
the population P. The  P (population) is a unary predicate with P(x)=TRUE 
indicating that the object x($\in \Omega$, that is element of a universe of 
objects) belongs to the population under considerations. If P and P' are 
populations such that $\forall_x P'(x)\rightarrow P(x)$ (that is membership 
in P' implies membership in P, or in other words: P' is a subpopulation of 
P), then we distinguish two cases:\\
case 1: $(\Prob{x}{P}P'(x))=0$ (that is probability of membership in P' with 
respect to P is equal 0) - then (according to \cite{Morgan:91} for any 
expression $\alpha(x)$ in free variable x the following holds for the 
population P': $(\Prob{x}{P'}\alpha(x))=1$\\
case 2: $(\Prob{x}{P}P'(x))>0$ then (according to \cite{Morgan:91} for any 
expression $\alpha(x)$ in free variable $x$ the following holds for the 
population P': 
$$(\Prob{x}{P'}\alpha(x))=  \frac {\Prob{x}{P}(\alpha(x) \land P'(x))}
                                {\Prob{x}{P}P'(x)}$$  
\begin{df} \label{MDef}
$X$ be a set-valued attribute taking as its values non-empty
subsets of a finite domain $\Xi$.
By a measurement method 
of value of the attribute $X$
we understand a function:
 $$M: \Omega \times 2^\Xi \rightarrow \{TRUE,FALSE\}$$ 
where $\Omega$ is the set of objects, (or population of objects)
such that 
\begin{itemize}
\item
 $ \forall_{\omega; \omega \in \Omega} \quad
M(\omega,\Xi)=TRUE$ (X takes at least one of values from $\Xi$)
\item
 $ \forall_{\omega; \omega \in \Omega} \quad
M(\omega,\emptyset)=FALSE$ 
\item 
 whenever 
$M(\omega,A)=TRUE$
for $\omega \in \Omega$, $A \subseteq \Xi$
 then for any $B$ such that $A \subset B$ $M(\omega,B)=TRUE$   
holds,
 \item 
 whenever 
$M(\omega,A)=TRUE$
for $\omega \in \Omega$, $A \subseteq \Xi$ and if $card(A)>1$ then there 
exists  $B$, $B \subset A$ such that $M(\omega,B)=TRUE$ holds.
\item 
for every $\omega$ and every $A$
either  
$M(\omega,A)=TRUE$  or 
 $M(\omega,A)=FALSE$ (but never both).
 \end{itemize}
$M(\omega,A)$  tells us whether or not any of the elements of the set A 
belong to the actual value of the attribute $X$ for the object $\omega$.
 \end{df}
\begin{df}
A {\em label} $L$ of an object $\omega \in \Omega$ is a subset of the
domain $\Xi$ of the attribute $X$. \\
A {\em labeling}  under the measurement method $M$  is a function $l: \Omega 
\rightarrow 2^\Xi$ such that for any object  $\omega \in \Omega$ either
$l(\omega)=\emptyset$ or $M(\omega,l(\omega))=TRUE$.\\
Each {\em labelled object}  (under the labeling $l$) 
consists of a 
pair $(\omega_j,L_j)$, $\omega_j$ - the j$^{th}$ object, $L_j=l(\omega_j)$ -
its label.\\
By a {\em population  under the labeling $l$} we understand the predicate
\linebreak 
$P:\Omega \rightarrow \{TRUE,FALSE\}$ of the form 
$P(\omega)=TRUE \  iff \ l(\omega) \neq \emptyset$
(or alternatively, the set of objects  for which this predicate is true) \\
 If for every  object of the 
population the label is equal 
 to $\Xi$ then  we  talk  of  an  {\em unlabeled  population} (under the 
labeling $l$), otherwise of a {\em pre-labelled} one.
\end{df}
\begin{df} 
Let $l$ be a labeling under the measurement method $M$. 
Let us consider the population under this labeling.
The modified measurement method 
$$M_l:
 \Omega \times 2^\Xi \rightarrow 
\{TRUE,FALSE\}$$
where $\Omega$ is the set of objects, 
is is defined as  
$$M_l(\omega,A)= M(\omega,A \cap l(\omega) )$$  
(Notice that 
$M_l(\omega,A)=FALSE$ whenever $A \cap l(\omega)= \emptyset$.)
\end{df}
\begin{df}
$$Bel_P ^{M}(A)=\Prob{\omega}{P}(\lnot M(\omega,\Xi-A))$$
which is the probability that the test M, while being true for A, rejects 
every hypothesis of the form X=$v_i$ for every  $v_i$  not  in  A 
for the population P.
We shall call this function "the belief exactly in  the result 
of measurement". 
\end{df}
\begin{df}
$$Pl_P ^{M}(A)=\Prob{\omega}{P}(  M(\omega,A))$$
which  is  the  probability  of  the  test  M  holding  for  
 A for the population P. Let us refer to this function as  the 
"Plausibility of taking any value from the set A".
\end{df}
\begin{df}
$$m_P ^{M}(A)=\Prob{\omega}{P}( \bigwedge_{B;B=\{v_i\}\subseteq A} M(\omega,B)
  \land \bigwedge_{B;B=\{v_i\}\subseteq \Xi-A} \lnot M(\omega,B) ) $$
which is the probability that all the  tests  for  the  singleton 
subsets of A are true and those outside of A are  false  for  the 
population P.
\end{df}
\begin{th} 
$m_P ^{M}$ is the mass Function in the sense of 
DS-Theory.
\end{th}
\begin{th} 
$Bel_P ^{M}$ is a Belief Function in the sense of 
DS-Theory   corresponding  to the  $m_P ^{M}$. 
\end{th}
\begin{th} 
$Pl_P ^{M}$ is a Plausibility Function in the sense of 
DS-Theory and it is the Plausibility  Function  corresponding  to 
the  $Bel_P ^{M}$. 
\end{th} %
\begin{df}
Let P be a population and $l$ its labeling. Then 

$$Bel_P    ^{M_l}(A)=\Prob{\omega}{P} \lnot M_l(\omega,\Xi-A)$$

$$Pl_P ^{M_l}(A)=\Prob{\omega}{P} M_l(\omega,A)$$

$$m_P ^{M_l}(A)=\Prob{\omega}{P} (\bigwedge_{B;B=\{v_i\}\subseteq A}
 M_l(\omega,B)
  \land \bigwedge_{B;B=\{v_i\}\subseteq \Xi-A} \lnot
 M_l(\omega,B))$$
\end{df}
\begin{th} 
$m_P ^{M_l}$ is the mass Function in the sense of 
DS-Theory.
\end{th}
\begin{th} 
$Bel_P ^{M_l}$ is a Belief Function in the sense of 
DS-Theory   corresponding  to the  $m_P ^{M_l}$. 
\end{th}
\begin{th} 
 $Pl_P ^{M_l}$ is a Plausibility Function in the sense of 
DS-Theory and it is the Plausibility  Function  corresponding  to 
the  $Bel_P ^{M_l}$. 
\end{th}
\begin{df}
Let $M$ be a measurement method, $l$ be a labeling under this measurement
method, and P be a population under this labeling (Note that the population
may also be unlabeled).
The  {\em (simple) labelling  process}   on    
the
population P 
is defined as a functional 
$\LP: 2^\Xi \times \Gamma \rightarrow \Gamma$, where $\Gamma$ is the set of  
all  possible labelings under $M$, 
such that for the given labeling $l$ and a given nonempty
set of attribute values $L$ ($L  \subseteq \Xi$), 
it delivers a new labeling $l'$ ($l'=\LP(L,l)$) such that for every object
$\omega \in \Omega$: 

1. if  $M_l(\omega,L)=FALSE$ then  
$l'(\omega)=\emptyset$\\
(that is l' discards a
labeled 
 object $(\omega,l(\omega))$ if $M_l(\omega,L )=FALSE$ 

2. otherwise $l'(\omega)=l(\omega) \cap L $
(that is l' labels the object with $l(\omega) \cap L $ otherwise.
\end{df}
\begin{df} "labelling  process  function" 
$m ^{\LP;L }: 2 ^\Xi \rightarrow [0,1]$:
 is defined as:
 $$m ^{\LP;L }(L )=1$$  
$$\forall_{B;  B  \in  2^\Xi,B \ne L } m ^{\LP;L }(B)=0$$
\end{df}

\begin{th} 
 $m ^{\LP;L }$ is a Mass Function in sense of DS-Theory.
\end{th}
\begin{th} 
\label{thSimpleLab}
Let $M$ be a measurement function, $l$ a labeling, P a population under
this labeling. Let $L $ be a subset of $\Xi$. 
Let $\LP$ be a labeling process and let $l'=\LP(L ,l)$.
Let P' be a population under the labeling $l'$.
Then 
 $Bel_{P'} ^{M_{l'}}$ is a  combination  via  DS  Combination 
rule of  $Bel ^{M_l}$,  and $Bel ^{\LP;L }$., that is:
$$Bel_{P'} ^{M_{l'}} = Bel_P ^{M_l} \oplus Bel ^{\LP;L }$$
\end{th}
\begin{df}
Let $M$ be a measurement method, $l$ be a labeling under this measurement
method, and P be a population under this labeling (Note that the population
may also be unlabeled).
Let  us  take  a  set  of (not  necessarily  disjoint) nonempty sets  of  
attribute values $\{L ^1, L ^2, ...,L ^k\}$    and 
let us define the  probability of selection as a function
$m ^{\LP, L ^1, L ^2, ...,L ^k}: 2 ^\Xi \rightarrow [0,1]$ such that
$$\sum_{A;A \subseteq \Xi}m ^{\LP, L ^1, L ^2, ...,L ^k}(A)=1$$
$$\forall_{A; A \in \{ L ^1, L ^2, ...,L ^k\}} 
m ^{\LP, L ^1, L ^2, ...,L ^k}(A)>0$$
$$\forall_{A; A \not\in \{ L ^1, L ^2, ...,L ^k\}} 
m ^{\LP, L ^1, L ^2, ...,L ^k}(A)=0$$
 The  {\em (general) labelling  process}   on    
the
population P 
is defined as a (randomized) functional 
$\LP: 2^{2^\Xi} \times \Delta
\times  \Gamma \rightarrow \Gamma$, where $\Gamma$ is the set 
of all  possible labelings under $M$, and $\Delta$ is 
a set of all possible probability of selection functions,
such that for the given labeling $l$ and a given 
 set  of (not  necessarily  disjoint) nonempty sets  of  
attribute values $\{L ^1, L ^2, ...,L ^k\}$    and 
a given probability of selection 
$m ^{\LP, L ^1, L ^2, ...,L ^k}$
it delivers a new labeling $l"$ such that for every object
$\omega \in \Omega$:

1. a label L, element of the set $\{ L ^1, L ^2, ...,L ^k\}$ 
is sampled randomly according to the probability distribution 
$m ^{\LP, L ^1, L ^2, ...,L ^k}$;
This sampling is done independently for each individual object,

2. if  $M_l(\omega,L)=FALSE$ then  
$l"(\omega)=\emptyset$\\
(that is l" discards an object $(\omega,l(\omega))$ if 
$M_l(\omega,L )=FALSE$ 

3. otherwise $l"(\omega)=l(\omega) \cap L $
(that is l" labels the object with $l(\omega) \cap L $ otherwise.)
\end{df}
\begin{th} 
$m ^{\LP,L ^1,...,L ^k}$ is a Mass Function in sense of DS-Theory.
\end{th}
\begin{th} 
Let $M$ be a measurement function, $l$ a labeling, P a population under
this labeling. 
Let $\LP$ be a generalized labeling process and let $l"$
be the result of application of the $\LP$ for the set
of labels from the set $\{ L ^1, L ^2, ...,L ^k\}$ 
 sampled randomly according to the probability distribution 
$m ^{\LP, L ^1, L ^2, ...,L ^k}$;.
Let P" be a population under the labeling $l"$.
Then 
The expected value 
over the set of all possible resultant labelings $l"$ (and hence
populations P") 
(or, more precisely, value vector) of 
$Bel_{P"} ^{M_{l"}}$ is a  combination  via  DS  Combination 
rule of  $Bel_P ^{M_l}$,  and $Bel ^{\LP,L ^1,...,L ^k}$., that is:
$$E(Bel_{P"} ^{M_l'}) = Bel_P ^{M_l} \oplus Bel ^{\LP,L ^1,...,L ^k}$$
\end{th}

}

\end{document}